\documentclass[letterpaper, 10 pt, conference]{ieeeconf}  

\IEEEoverridecommandlockouts                              

\usepackage{graphics} 
\usepackage{epsfig} 
\usepackage{amsmath} 
\usepackage{amssymb}  
\usepackage{subfigure}
\usepackage{amsfonts}

\usepackage{amsthm}
\usepackage{gensymb}
\usepackage[utf8]{inputenc}
\usepackage[english]{babel}
\usepackage{color}
\usepackage{mathtools}
\usepackage{algorithm}
\usepackage[noend]{algpseudocode}
\usepackage[font=footnotesize]{caption}
\usepackage{tabularx}
\usepackage{bm}
\usepackage{multirow}
\usepackage{xcolor}
\usepackage{cite}
\usepackage{lipsum}

\include{shortcuts}

\newcommand{\fig}[1]{Fig.~\ref{#1}}

\usepackage[normalem]{ulem}

\title{\LARGE \bf
Two-Stage Clustering of Human Preferences\\ for Action Prediction in Assembly Tasks\vspace{-1.0 ex}}

\author{Heramb Nemlekar, Jignesh Modi, Satyandra K. Gupta and Stefanos Nikolaidis
\thanks{Heramb Nemlekar, Jignesh Modi, Satyandra K. Gupta and Stefanos Nikolaidis are with the University of Southern California, USA.
        {\tt\small \{nemlekar, jigneshm, guptask, nikolaid\}@usc.edu}}%
\vspace{-0.5 ex}}

\begin{document}

\maketitle
\thispagestyle{empty}
\pagestyle{empty}

\begin{abstract}
To effectively assist human workers in assembly tasks a robot must proactively offer support by inferring their preferences in sequencing the task actions. Previous work has focused on learning the dominant preferences of human workers for simple tasks largely based on their intended goal. However, people may have preferences at different resolutions: they may share the same high-level preference for the order of the sub-tasks but differ in the sequence of individual actions. We propose a two-stage approach for learning and inferring the preferences of human operators based on the sequence of sub-tasks and actions. We conduct an IKEA assembly study and demonstrate how our approach is able to learn the dominant preferences in a complex task. We show that our approach improves the prediction of human actions through cross-validation. Lastly we show that our two-stage approach improves the efficiency of task execution in an online experiment and demonstrate its applicability in a real-world robot-assisted IKEA assembly.
\end{abstract}


\section{Introduction}
There are many assembly, service, repair, installation, and construction applications where many different workers may need to perform the same task. For example, consider the task of replacing a bearing on a machine tool located at the shop floor. Even though these tasks have fixed guidelines, there is some variability in the way each worker performs a task because of individual preferences. Robots can help in improving the efficiency of such tasks by adapting to the individualized preferences of human workers and proactively supporting them in their task. 

While the space of possible preferences can be very large,  previous work has shown that people can be grouped to a few ``dominant'' preferences: In human-agent teams, users cluster to a set of ``reasonable'' behaviors, where people in the same cluster have similar beliefs~\cite{pynadath2018clustering}. Similar groupings exist in game playing~\cite{ramirez2010player,tondello2017elements,thawonmas2007detection,holmgaard2013decision} and education~\cite{peckham2012mining,merceron2004clustering,thuillier2017clustering,furletti2012identifying}. 

\begin{figure}[t!]
\centering
\subfigure[User preference 1]{%
\includegraphics[width=0.475\linewidth]{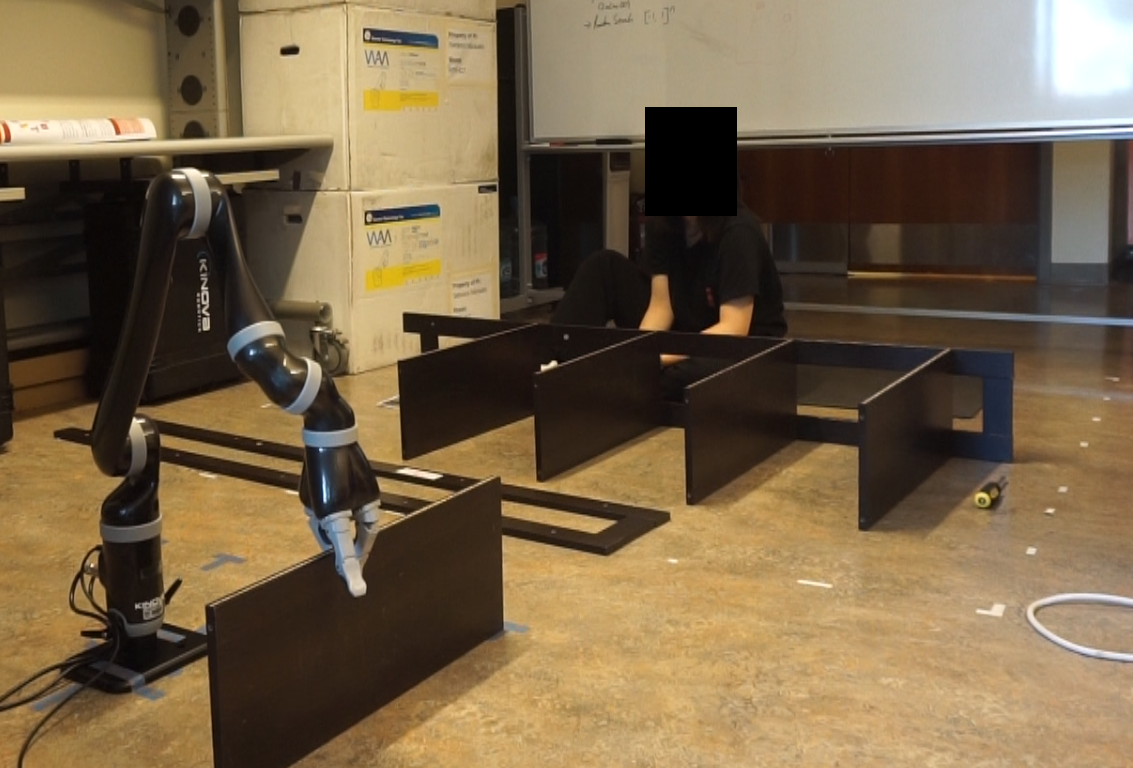}\vspace{-0.5ex}
\label{fig:user1}}%
\subfigure[User preference 2]{%
\includegraphics[width=0.475\linewidth]{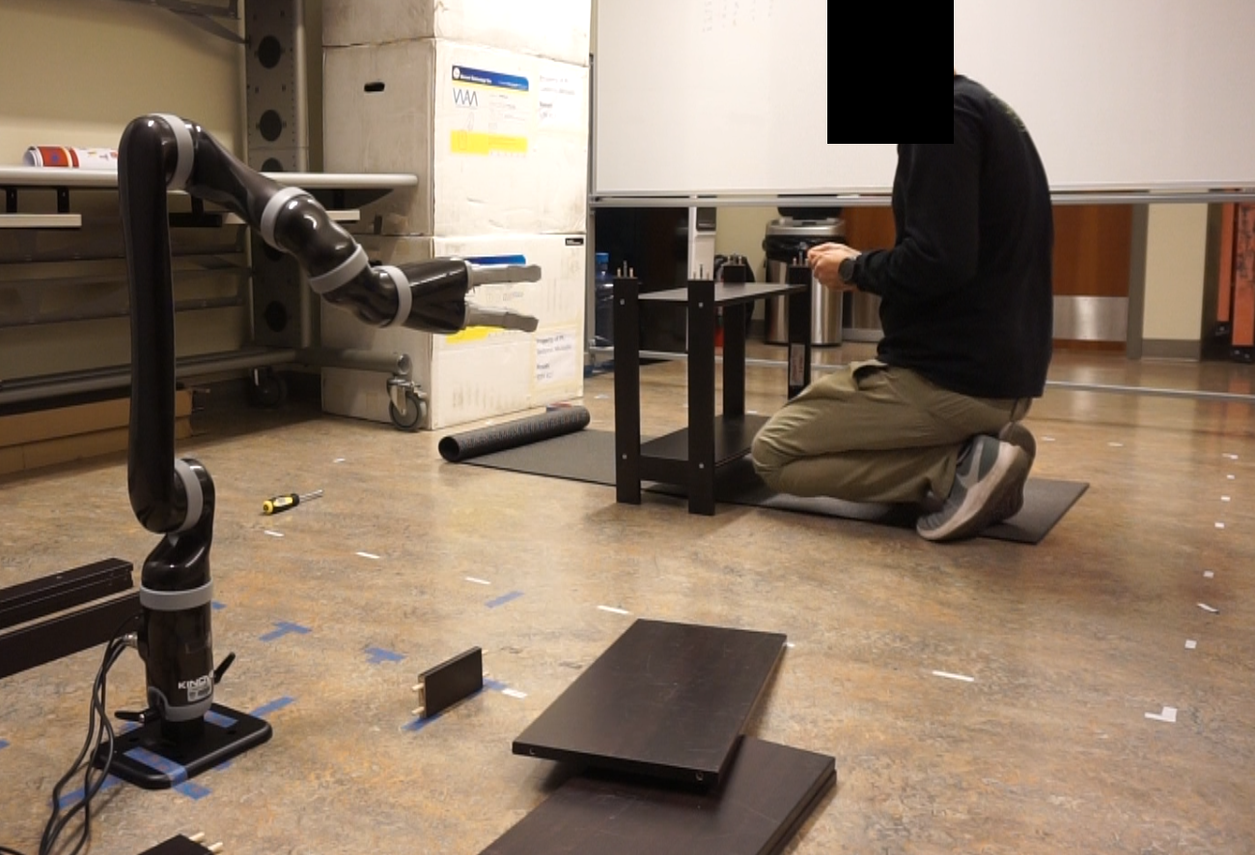}\vspace{-0.5ex}
\label{fig:user2}}%
\vspace{-1.5ex}
\caption{Robot-assisted IKEA assembly. (a) Some users preferred to connect all the shelves in a row. (b) Other users connected just two shelves to the small boards first. For effectively assisting the users, a robot must predict their preference and supply the parts accordingly.}
\label{fig:user_preferences}
\vspace{-3.5ex}
\end{figure}

What makes the problem particularly challenging in complex tasks, is that these dominant preferences exist in \textit{different resolutions}. In an IKEA assembly study that we use as a proof-of-concept throughout the paper, we observed that some participants preferred to assemble all shelves in a row, while others alternated between assembling the shelves and assembling the boards (Fig.~\ref{fig:user_preferences}). Moreover, within the first group, participants also differed in how they connected the shelves to the boards: some connected all shelves to boards on just one side first (as in Fig.~\ref{fig:user1}), while others preferred to connect each shelf to boards on both sides. 

 There are two scenarios where learning the dominant preferences can help. First, a new worker may be asked to perform the same task that we have collected demonstrations for. A robotic assistant would need to infer the preferences of the new worker at different resolutions by associating them with previously learned dominant preferences, and proactively assist them by anticipating their next action. Second, a worker that demonstrated task A may be assigned to work on a slightly different task B, e.g., assemble an IKEA bookcase of different shape. While the task has changed, e.g., the number and type of shelves, some action sequences may be shared between A and B, e.g., connecting shelves to the boards. The robot should be able to use its knowledge about the worker's preference when assisting them in task B. 

Both these scenarios require \textit{learning preferences at different levels of abstraction}. We propose abstracting action sequences to sequences of \textit{events}, which are transferrable units shared among different tasks and workers. A high-level preference is captured by a sequence of events, while a lower-level preference captures how each event is executed. Using insights from clickstream analysis~\cite{antonellis2009algorithms}, we propose clustering users on two levels, over events and over actions within each event, as opposed to clustering based on the sequences of individual actions as in previous work~\cite{nikolaidis2015efficient}.

We show the applicability of our method on the first scenario, where a new worker executes the same task that we have demonstrations for. If the task has $n$ dominant preferences at all levels, then we need to observe $n \times m$ workers. $n$ is a number usually between three to five~\cite{nikolaidis2015efficient}, while larger values of $m$ allow for more robust inference, since there may be variability in the exact sequence of actions of the workers with the same preference. 
We conduct a user study where 20 users assemble an IKEA bookcase, and we show that we can learn the high and low-level preferences of the users, which enables accurate prediction for a new user performing the same task. Through an online assembly experiment we show that assisting users this way improves task efficiency. We finally show the applicability of the system in a real-world robot-assisted IKEA assembly demo.

\section{Related Work}\label{sec:related}

\textbf{Modeling human preferences.} 
User preferences for collaborative tasks are often measured through surveys  \cite{erdogan2017effect, carmichael2020human}. Preferences can then be modelled as a feed-forward neural network that maps task metrics to the survey responses of users \cite{erdogan2017prediction}. Human preferences for robot actions can also be measured from EEG signals \cite{iwane2019inferring}, or from a short window of human arm motion \cite{butepage2017anticipating,schmerling2018multimodal,park2017intention} or gaze pattern \cite{huang2016anticipatory}.
Past work has also focused on incorporating user preferences in task assignment and scheduling, where the user preferences are included as a constraint \cite{gombolay2015coordination, wilcox2013optimization} or in the objective function \cite{gombolay2017computational} of the scheduling problem. 

For task planning, like in our approach, preference is considered as the subset of action sequences from the set of multiple sequences that solve the same task \cite{munzer2017preference}. Related work includes learning user preferences during an assembly task from demonstrations\cite{argall2009survey, ravichandar2020recent}, via interactive reinforcement learning \cite{akkaladevi2016towards, munzer2017preference} or active reward learning \cite{sadigh2017active, biyik2018batch, biyik2020active}, where previous demonstrations can be used as priors \cite{palan2019learning, biyik2020learning}. Human feedback can also be used to directly modify the policy instead of the reward function \cite{griffith2013policy}. 


\textbf{Clustering dominant preferences.} While each user can have a different preference, our goal is to cluster the users to a small set of \textit{dominant preferences}. Such groups of similar users, also called personas, can be built manually from questionnaires \cite{madureira2014using} or through collaborative filtering \cite{zhou2008large, abdo2016organizing}. Other related work includes identifying different driver styles~\cite{sundbom2013online,wang2017driving,higgs2013two} using features from vehicle trajectories, human motion prototypes~\cite{luber2012socially} for robot navigation and human preference stereotypes for human-robot interaction~\cite{wagner2012using}. 

Most relevant to ours is prior work in identifying dominant user preferences from sequences of user actions in a surface refinishing task \cite{nikolaidis2015efficient}. Users with similar action transition matrices were clustered using a hard Expectation Maximization (EM) algorithm to obtain the dominant clusters. However the task was simple and thus one-stage clustering of the transition matrices was able to capture the user preferences which were largely encoded in the final position of the robot.

\textbf{Clustering sequences.} The problem of grouping users based on their preferred sequence of doing tasks is similar to the problem of \textit{clickstream analysis} \cite{banerjee2001clickstream, antonellis2009algorithms, wang2016unsupervised, wang2017clickstream}. A clickstream is a sequence of timestamped events generated by user actions (clicking or typing) on a webpage and hence is comparable to a sequence of actions. Prior work clusters clickstreams of multiple users based on their longest common sub-sequence \cite{banerjee2001clickstream} or frequency of sub-sequences \cite{wang2017clickstream}. 

To cluster users at different resolutions, prior work uses Levenshtein distance to form macro and micro preference clusters \cite{antonellis2009algorithms}. Recent work uses a similarity graph \cite{wang2016unsupervised} where similarity between users is measured by comparing the sub-sequences of their clickstreams. Hierarchical clustering is used to partition the similarity graph into high-level clusters which are then further partitioned based on features that were unused for the high-level clustering. We bring these insights from clickstream clustering to the robotics problem of clustering the action sequences of users in assembly tasks.



\section{Methodology}\label{sec:method}

\begin{figure}[hbt!]
\vspace{-1ex}
\centering
\includegraphics[width=0.95\linewidth]{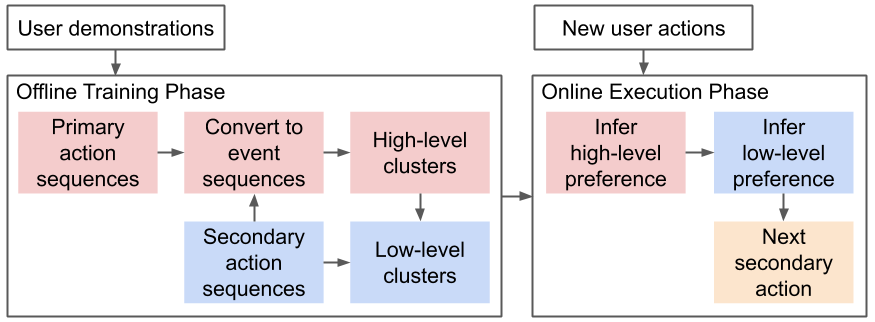}\vspace{-0.5 ex}
\caption{Flowchart of our proposed two-stage clustering and inference method}
\label{fig:flowchart}
\vspace{-1.5 ex}
\end{figure}

The proposed method consists of two phases (see \fig{fig:flowchart}): (1) an offline training phase which takes as input a set of user demonstrations of the entire assembly task and learns the dominant preference clusters at different resolutions, and (2) an online execution phase where we estimate the probability of a new user belonging to one of the clusters based on their observed actions, and predict the next robot action.

Based on our observation that users prefer to perform actions that require the same parts in a row; we first convert each user demonstration into a sequence of such $events$. Thus each event in a demonstration requires a specific set of parts to be supplied by the robot i.e. a specific set of \textit{secondary actions} (non-critical actions like supplying parts).
The \textit{high-level preference} of each user is thus the order in which they perform the events. Further for each event, the users may have a different \textit{low-level preference} of the order in which the set of parts should be supplied i.e. order in which the secondary actions must be performed.

We learn the high and low-level preferences in the offline phase by clustering users based on their sequence of events and sequence of secondary actions respectively. Accordingly in the online execution phase we first infer the high-level preference of a new user and then infer the low-level preference to determine the next secondary action to execute.


\section{Offline Training Phase}\label{sec:offline}
We assume a set of demonstrated action sequences $X$, with one $x \in X$ per user. Similar to prior work \cite{grigore2018preference}, we distinguish the actions $A$ in the demonstrated sequences into two types: \textbf{primary actions} ($a^p \in A^P$) which are the task actions that \textit{must be performed by the user}, and \textbf{secondary actions} ($a^s \in A^S$) which are the supporting actions that \textit{can be delegated to the robot}. 
For instance, a primary action is connecting a shelf, while a secondary action is bringing the shelf to the user. Therefore, $A = \{A^{P}, A^{S}\}$.

In the training phase, each user demonstration $x$ is some sequence of primary and secondary actions. For example, demonstration $[a_{1}^{s}, a_{2}^{s}, a_{1}^{p}, a_{2}^{p}, \ldots, a_{M}^{s}, a_{N}^{p}]$ has $M$ secondary and $N$ primary actions. 
We wish to model the online execution as a turn-taking model where at each time step $t$, a set of secondary action $s_t$ is performed, followed by a primary action $a_{t}^{p}$. We choose this model as at each time step $t$, we want the robot to predict the next primary action ($a_{t+1}^{p}$) of the user and proactively perform the set of secondary actions ($s_{t+1}$) that comes before it. Thus, we group consecutive secondary actions into a set of secondary actions $s$ and insert a $NOOP$ (no-operation) action between consecutive primary actions to obtain, $x = [s_{1}, a_{1}^{p}, s_{2}, a_{2}^{p}, \ldots, s_{N}, a_{N}^{p}]$. Where in this example, $s_{1} = [a_{1}^{s}, a_{2}^{s}]$ is the set of secondary actions that must be executed before the primary action $a_{1}^{p}$, while $s_{2} = [NOOP]$ means that no other secondary action is required to be executed before $a_{2}^{p}$.


\subsection{Converting User Demonstrations to Event Sequences} \label{subsec:converting}

We first convert each user demonstration to a sequence of \textit{events}. An event $e$ is defined as - consecutive primary actions that require the same set of secondary actions. Thus an event from time step $t_{a}$ to $t_b$ is the tuple $e_{t_{a} : t_{b}} = (p_{t_{a} : t_{b}}, s_{t_{a} : t_{b}})$, where $p_{t_{a} : t_{b}}$ is the sequence of primary actions $[a_{t_a}^{p}, \ldots, a_{t_b}^{p}]$ with similar preceding secondary actions $s_{t_{a} : t_{b}} = [s_{t_a}, \ldots, s_{t_b}]$. Two events are equal if they share the same \textit{set of secondary actions} - $\{s_{t_a}, \ldots, s_{t_b}\}$.

Consider the demonstration, 
$x = [[a_{1}^{s}, a_{2}^{s}], a_{1}^{p}, [NOOP],$  $a_{2}^{p}, [a_{3}^{s}], a_{3}^{p}]$.
We first set the primary sequence $p_{1:1} = [a_{1}^{p}]$ (Line \ref{step:primary} of Algorithm \ref{algo:events}) and corresponding secondary sequence $s_{1:1} = [[a_{1}^{s}, a_{2}^{s}]]$ (Line \ref{step:secondary}) for event $e_{1:1}$. Then we append the next primary action $a_{2}^{p}$ to the same event if the secondary action preceding it ($s_2$) is similar to any of the secondary actions in $s_{1:1}$ i.e. $s_2 \subseteq \{s_{1:1}\}$ (Line \ref{step:cond}). As $[NOOP]$ is no operation (empty set), we consider it to be a subset of every $\{s_{t_{a}:t_{b}}\}$. Therefore, $p_{1:2} = [a_{1}^{p}, a_{2}^{p}]$ and $s_{1:2} = [[a_{1}^{s}, a_{2}^{s}], [NOOP]]$. Similarly, if the secondary action $[a_{3}^{s}]$ preceding the next primary action $a_{3}^{p}$ is a subset of $\{s_{1:2}\}$, all primary actions belong to the same event $e_{1:3}$, with $p_{1:3} = [a_{1}^{p}, a_{2}^{p}, a_{3}^{p}]$ and $s_{1:3} = [[a_{1}^{s}, a_{2}^{s}], [NOOP], [a_{3}^{s}]]$. In this case, the event sequence is $x^e = [e_{1:3}]$. 

However if $a_{3}^{s} \not\in [a_{1}^{s}, a_{2}^{s}]$, then $a_{3}^{p}$ will belong to a new event $e_{3:3}$, with $p_{3:3} = [a_{3}^{p}]$ and $s_{3:3} = [a_{3}^{s}]$. Thus, in this case, the event sequence will be $x^e = [e_{1:2}, e_{3:3}]$ where $e_{1:2} = (p_{1:2}, s_{1:2})$. This process is described in Algorithm~\ref{algo:events}.


\begin{algorithm}
\caption{Convert demonstration $x$ to event sequence $x^e$}
{\fontsize{10}{11}\selectfont
\begin{algorithmic}[1]
\Require Demonstration $x$, $A^{P}$, $A^{S}$
\State \textbf{set} $p = [\;]$, $s = [\;]$, $x^{e} = [\;]$
\For{$x[i]$ \textbf{in} $x$}
    \If{$x[i] \in A^P$}
        \State \textbf{append} $p \leftarrow [p, \; x[i]]$ \label{step:primary}
    \EndIf
    \If{$x[i] \in A^S$}
        \If{$s \neq [\;]$ \textbf{and} $x[i] \notin s$}\label{step:cond}
            \State \textbf{set} $e = (p, s)$
            \State \textbf{append} $x^{e} \leftarrow [x^{e}, \;e]$ \label{step:eseq}
            \State \textbf{set} $p = [\;]$, $s = [\;]$, $x^{e} = [\;]$ \label{step:nexte}
        \EndIf
        \State \textbf{append} $s \leftarrow [s, \;x[i]]$ \label{step:secondary}
    \EndIf
\EndFor
\State \Return $x^e$
\end{algorithmic}
}
\label{algo:events}
\end{algorithm}

\vspace{-1ex}
\subsection{Two-Stage Clustering}

\subsubsection{\textbf{Clustering Event Sequences}} Once we have obtained an event sequence $x^e$ for each user, we cluster the event sequences to determine the \textit{dominant high-level clusters} $z_h \in Z_H$. Details of the method for clustering the sequences are provided in Sec.~\ref{subsec:clustering-method}.

\subsubsection{\textbf{Clustering Secondary Action Sequences}} To learn the low-level preferences for each event $e$, 
we cluster participants based on the sequence of secondary actions of the event to determine the \textit{dominant low-level clusters} $z_{l} \in Z_{L}^{e}$. We use secondary actions since, from a robot's perspective, primary actions that require the same assistive response (i.e. preceding secondary actions) can be considered identical. 



\begin{figure}[hbt!]
\centering
\subfigure[$e_{boards}$ is $1^{st}$ event for user A]{%
\includegraphics[width=0.485\linewidth]{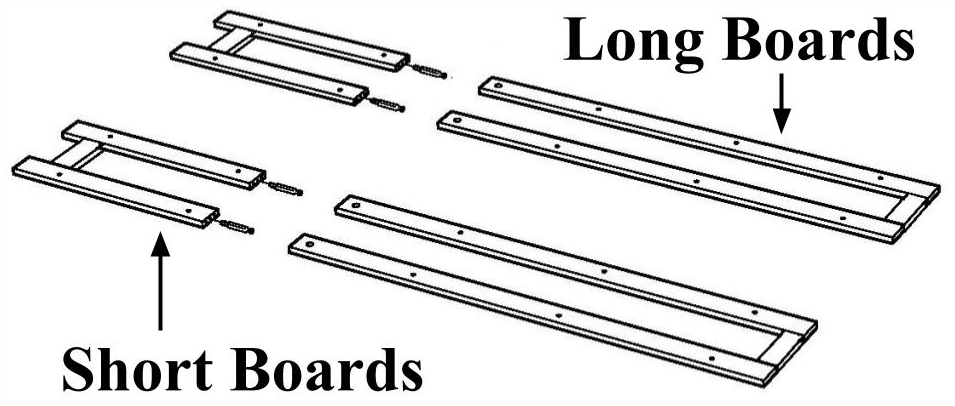}
\label{fig:high1}}%
\subfigure[$e_{boards}$ is $2^{nd}$ event for B and C]{%
\includegraphics[width=0.485\linewidth]{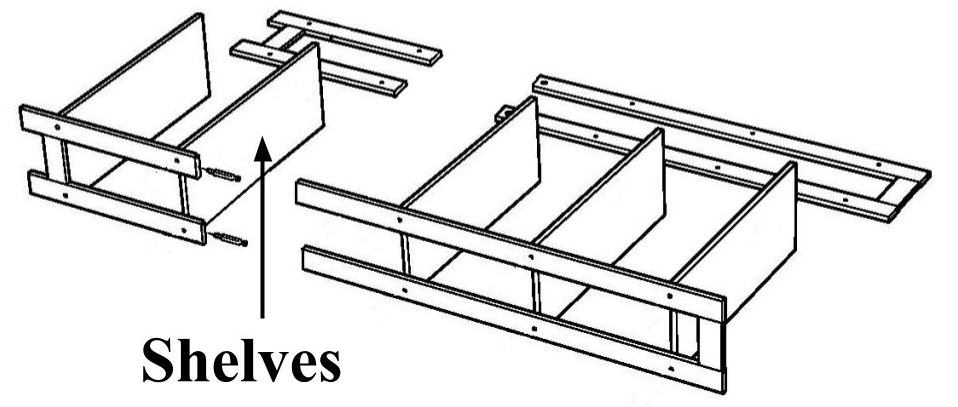}
\label{fig:high2}}%
\vspace{-1ex}
\caption{Visualization of event $e_{boards}$: connect long and short boards. For B and C, $e_{boards}$ is after event $e_{shelves}$: connect the shelves to the boards.}
\label{fig:high_level_visual}
\vspace{-1ex}
\end{figure}

\subsubsection{\textbf{Intuition}}
Consider the following example with two events $e_{boards}$ and $e_{shelves}$ (see \fig{fig:high_level_visual}), and three users with event sequences A: $[e_{boards},e_{shelves}]$, B: $[e_{shelves},e_{boards}]$, C: $[e_{shelves},e_{boards}]$. User A prefers to assemble the boards first and then connect shelves to the boards, while users B and C prefer to connect the shelves before assembling the boards. The high-level clustering assigns users B and C to the same cluster, different than A. Here, the set of secondary actions to execute $e_{shelves}$ for users B and C, is different to that for user A. B and C need both the shelf and the board to be supplied as it is the first event, while A only needs the shelves, as the boards were previously supplied for $e_{boards}$. 

Now assume that B performs $e_{shelves}$ by connecting each shelf to the boards on only one side as in \fig{fig:high2}, while C connects each shelf to boards on both sides (not shown). In this case, even though B and C share the same set of secondary actions for event $e_{shelves}$, the sequence of the actions would be different. The robot will perform one $NOOP$ after supplying a shelf to B as it waits for B to connect the shelf to the two boards on one side; whereas for C, the robot will perform three $NOOP$s after supplying the shelf as C connects that shelf to the two boards on both sides. Therefore the low-level clustering will place B and C in different low-level clusters.

In summary, \textit{first stage clustering enables learning the set of secondary actions required, while second stage specifies the order these actions should be performed}. On the other hand, one-stage clustering as in prior work\cite{nikolaidis2015efficient} would assign A, B and C in three different clusters, \textit{losing the information that B and C share the same set of secondary actions}!

\subsection{Clustering Method} \label{subsec:clustering-method} For each stage, we apply \textit{hierarchical clustering} \cite{bar2001fast,mullner2011modern} considering the distance metric: \textit{Levenshtein distance} \cite{navarro2001guided} ($d_{lev}$) used in clickstream analysis \cite{yujian2007normalized,antonellis2009algorithms} with a custom modification of the levenshtein distance ($d_{mod}$). We consider the operators: $add$ and $delete$ as in Levenshtein distance and instead of the $substitute$ operator we use the operation $shift_1$. $delete(i)$ removes the $i^{\text{th}}$ element of a sequence. $add(i)$ inserts an element into the $i^{\text{th}}$ position of a sequence given that it is empty. $shift_1(i)$ shifts the $i^{\text{th}}$ element to neighbouring the $i$$+$$1$ or $i$$-$$1$ position in the sequence given that it is empty. Each operation has a cost of $1$. The $shift_1$ operation allows us to consider sequences that only have two elements in swapped positions as closer than sequences that have two completely different elements in those positions.

The number of clusters depends on a distance threshold. We generate clusters for increasing distance thresholds, and select the optimal distance based on the variance ratio criterion (VRC) \cite{calinski1974dendrite} (also called \textit{calinski-harabasz score}) which is a common metric for distance-based clustering \cite{el2019evaluation}. VRC is the ratio of the sum of between-clusters dispersion and the sum of inter-cluster dispersion. Therefore for a high VRC the clusters are well separated from each other and the samples within the cluster are dense.

\section{Online Execution Phase}\label{sec:execution}

In the online execution phase we infer the high and low level preferences of new users as they are executing the task. At each time step $t$, as the user performs a primary action, the robot predicts the next secondary action. If the robot prediction is incorrect, the user performs the desired secondary action according to their preference.

\subsubsection{Inferring high-level preference}
At each time step $t$, we observe the primary action of a new user and append it to the actions observed so far. We then convert the current sequence of actions $x_{1:t}$ of the new user to a sequence of events $x_{1:t}^e$ in the same way as in the offline phase. We use Bayesian inference to predict the high-level preference by computing the probability of observing the event sequence $x_{1:t}^e$ for each high-level cluster $z_h \in Z_H$.
$$p(z_h | x_{1:t}^e) \propto p(x_{1:t}^e | z_h) p(z_h)$$
Here, $p(z_h)$ is simply the ratio of the number of users in the cluster $z_h$ to the total number of users in all the clusters $Z_{H}$.
However for calculating $p(x_{1:t}^e | z_h)$ we re-compute the event sequences of users in $z_h$ considering their action sequences only up to the time step $t$. Therefore:
$$p(x_{1:t}^e | z_h) = \frac{\text{No. of users in } z_h \text{ with same } x_{1:t}^e}{\text{Total no. users in } z_h}$$
We then determine the high-level preference as $z_{h}^* = \arg\max_{z_h \in Z_H} p(z_h | x_{1:t}^e)$. If there are two high-level clusters with the same maximum probability, we select one randomly.   


\subsubsection{Inferring low-level preference}
Once we infer the high-level preference $z_{h}^{*}$ of the new user, we identify the most likely event sequence $x^{e*}$ in that cluster. We assume that the new user follows that sequence $x^{e*}$ to index the event ongoing at time step $t+1$ i.e. $e_{t+1}$ (in a slight abuse of notation). Given the sequence of secondary actions $s^{e_{t+1}}$ performed so far within the event $e_{t+1}$, we use Bayesian inference to infer the low-level preference $z_{l} \in Z_{L}^{e_{t+1}}$, where $Z_{L}^{e_{t+1}}$ is the set of low-level preferences
for the event $e_{t+1}$: 
$$p(z_{l} | s^{e_{t+1}}) \propto p(s^{e_{t+1}} | z_{l}) p(z_{l})$$

We select the most likely low-level preference $z_l^*$, identically to the high-level preference case. The robot can then perform the most likely secondary action $s_{t+1}$ in $z_l^*$  to proactively assist the user. If the user accepts $s_{t+1}$ we append that to $x_{1:t}$. If the user rejects $s_{t+1}$ and performs a different secondary action $s_{t+1}^{'}$ instead, we append $s_{t+1}^{'}$ to $x_{1:t}$.

\section{User Study}\label{sec:study}

We wish to show that the proposed method can effectively identify the dominant preferences of users in an IKEA bookcase assembly task, and use the found preferences to accurately predict the next secondary action of a new user. 



\begin{figure}[hbt!]
\vspace{-0.5 ex}
\centering
\includegraphics[width=0.6\linewidth]{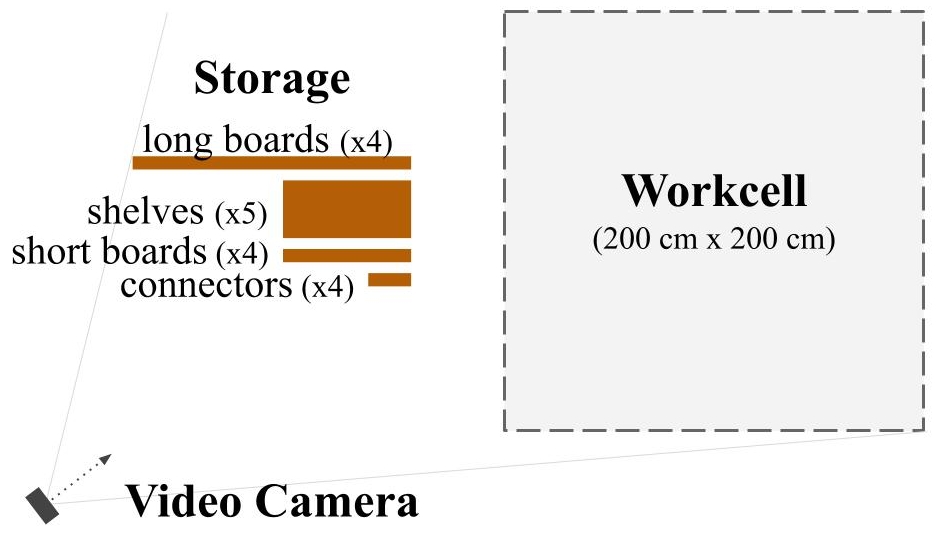}
\includegraphics[width=0.26\linewidth]{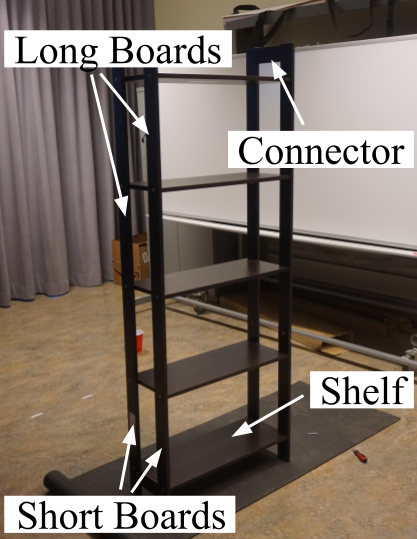}
\caption{Top view of study setup (left) and assembled bookcase (right)}\vspace{-0.5 ex}
\label{fig:study}
\vspace{-1.5 ex}
\end{figure}

\subsection{Study Setup}

We conducted a user study where subjects assembled an IKEA bookcase in a laboratory setting shown in Fig.~\ref{fig:study}-left. We divided the space into a storage area where all the parts were placed and a workcell where the user assembled the bookcase. We recruited $20$ subjects out of which $18$ ($M = 11$, $F = 7$) successfully completed the study. 

We provided each subject with a labelled image of the bookshelf (\fig{fig:study}-right) and demonstrated how the connections are made. Participants then practiced the connections for five minutes. We \textit{did not provide any instructions regarding the order of the assembly}. We informed participants that they had to assemble the shelf as fast as they can, and asked them to plan their preferred sequence beforehand, to ensure that their plan is well-thought.
The study was approved by the Institutional Review Board (IRB) of the University of Southern California. Participants were compensated with $10$ USD and the task lasted about one hour.  


\textit{Measurements:} We recorded a video of the users assembling the shelf. We annotated all participants' actions in the video using the video annotation tool ELAN \cite{wittenburg2006elan}. The annotation was done by 2 independent annotators who followed a common annotation guide.


\subsection{Analysis of User Preferences}

We considered bringing any part from the storage to the workcell as a secondary action and all connections in the assembly as primary actions. The bookcase had $4$ types of parts: long boards, short boards, connectors and shelves (total $17$ parts), and $32$ different connections. Thus each user demonstration was a sequence of $N = 32$ time steps.

\begin{figure}[ht]
\centering
\includegraphics[width=0.845\linewidth]{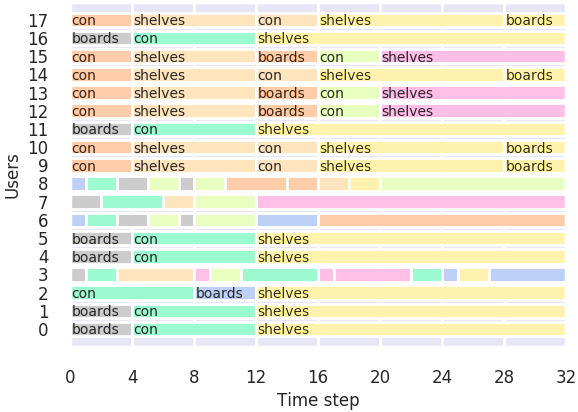}\vspace{-0.5ex}
\caption{Event sequences. `boards' refers to an event of connecting long and short boards, `con' refers to an event of connecting assembled boards using connectors, `shelves' refers to an event of connecting shelves.}
\label{fig:event_sequences}
\vspace{-2.5ex}
\end{figure}

\textbf{Event Sequences.} We first visualize the sequence of events for each user (shown in \fig{fig:event_sequences}). Users $[0, 1, 4, 5, 11, 16]$ had the same event sequence: short and long board connections (shown in grey), connector and board connections (green), and shelf and board connections (yellow). Similarly, other groups of users like $[12, 13, 15]$, and $[3, 9, 10, 14, 17]$ also had the same event sequences.

The assembly task is fairly complex; the $32$ primary actions can be ordered in more than  $24! $ ways! However most users preferred to perform similar actions in a row: $14$ users performed all long and short board connections in a row, and $7$ users connected all connectors and all shelves in a row. 

\begin{figure}[t!]
\centering
\subfigure[High-level clusters]{%
\includegraphics[width=0.49\linewidth]{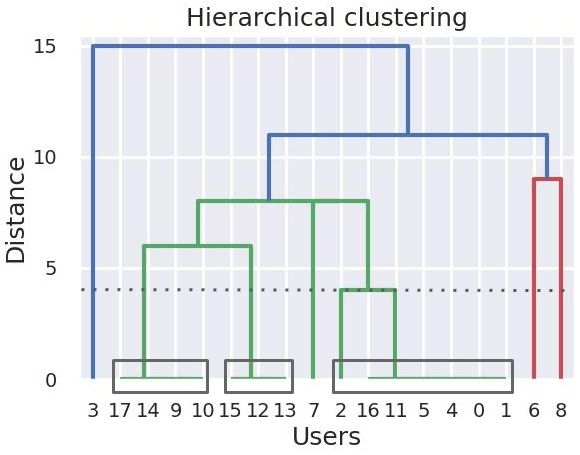}
\label{fig:high-clusters}}%
\subfigure[Low-level clusters]{%
\includegraphics[width=0.49\linewidth]{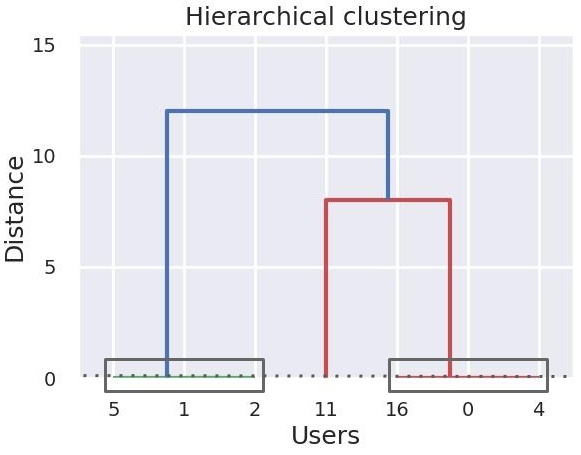}
\label{fig:low-clusters}}%
\vspace{-1ex}
\caption{IKEA assembly user study: (a) Dominant high-level clusters (grey rectangles) formed by clustering the event sequences. (b) Dominant low-level clusters (grey rectangles) formed by clustering the secondary actions sequences for the event of all shelf connections (shown in yellow in \fig{fig:event_sequences}).}
\vspace{-3ex}
\label{fig:offline_training}
\end{figure}

\textbf{Dominant clusters.} To find the high-level preferences, we cluster the event sequences using the modified Levenshtein distance metric (Sec.~\ref{subsec:clustering-method}) which results in the hierarchy shown in \fig{fig:high-clusters}. We partition the users at a distance threshold of $d_{mod} = 4$ (shown as dotted line) based on the VRC \cite{calinski1974dendrite} to obtain three dominant high-level clusters (shown in grey rectangles). Therefore we see that users cluster to a small set of dominant preferences despite not being provided any instructions regarding the order of assembly.

Users had different preferences for the same event as well.  For example, within the event of performing all shelf connections in a row, shown in yellow in \fig{fig:event_sequences}, clustering the sequences of secondary actions of all users results in the hierarchy shown in \fig{fig:low-clusters}. Partitioning at the optimal distance threshold $d_{mod} = 0$ gives us two dominant low-level clusters (shown in grey rectangles). Users 0, 4 and 16 prefer to connect each shelf to boards on one side before connecting on the other side, whereas users 5, 1 and 2 prefer to connect each shelf to boards on both sides at a time.

\subsection{Evaluation \& Results}\label{sec:eval}

We want to show that our two-stage clustering approach is better at predicting the next secondary action of the robot for a new user, as compared to a one-stage approach. We make the following hypothesis:

\textbf{H1.} Clustering users based on just their sequence of \textit{events} improves the accuracy of predicting the next secondary action, compared to clustering based on the sequences of individual primary actions.

\textbf{H2.} The two-stage clustering of users improves the accuracy of predicting the next secondary action, compared to clustering based on just the sequences of events. 

\textbf{Experiment details.} Using the demonstrated action sequences in the user study, we performed leave-one-out cross-validation, where we removed one participant and generated the preference clusters from the rest of the demonstrations. At each time step, the system was provided the corresponding primary action of the removed participant. We then inferred the high and low level preference of the participant based on the history of actions provided so far (Sec.~\ref{sec:execution}) and selected the secondary action that must be performed before the next primary action of the user. At the end, we compare the cross-validation accuracy of predicting the next secondary action at each time step, averaged over  100 trials for each new user.

\begin{figure}[hbt!]
\vspace{-0.5 ex}
\centering
\subfigure[Importance of high-level]{
\includegraphics[width=0.475\linewidth]{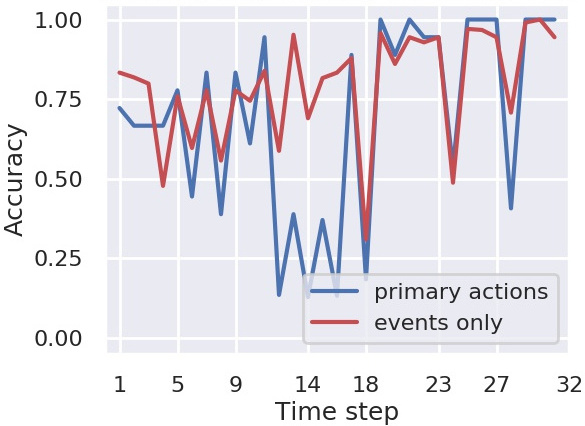}\label{fig:fullseq}}
\subfigure[Importance of low-level]{
\includegraphics[width=0.475\linewidth]{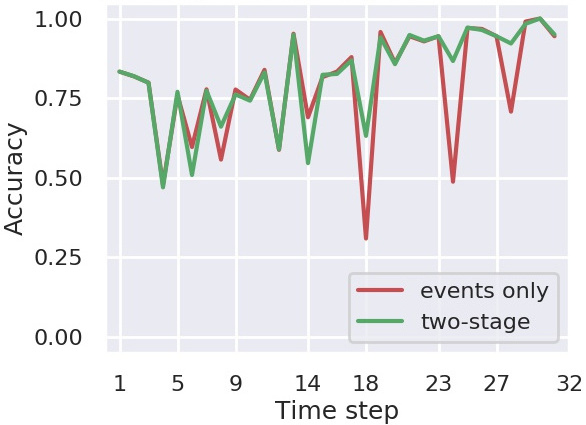}\label{fig:twostage}}
\vspace{-1ex}
\caption{Cross-validation accuracy for $18$ users averaged over $100$ trials. (a) Clustering on event sequences compared to clustering primary action sequences. (b) Two-stage clustering compared to clustering event sequences.}
\label{fig:performance}
\end{figure}

\subsubsection{Importance of High-level Clusters} We compare the prediction results from clustering based on sequence of events (without second-stage clustering) to clustering based on the sequence of primary actions. While clustering based on event sequences leads to three dominant clusters, clustering based on primary action sequences generates only two dominant clusters for a distance threshold of $63$ (based on VRC), resulting in high variability in the sequences in each cluster. A two-tailed paired t-test showed a statistically significant difference ($t(17) = -3.232$, $p = 0.004$) in prediction accuracy averaged over all timesteps and trials, between the event-based method ($M = 0.796,~SE = 0.041$) and the baseline ($M = 0.693,~SE = 0.041$). This result supports hypothesis \textbf{H1}.

\subsubsection{Importance of Low-level Clusters} We also compare the prediction accuracy with and without the second stage clustering, averaged over all timesteps and trials. A paired t-test showed a statistically significant difference ($t(17) = -2.34$, $p = 0.03$) between predicting with event sequences only ($M = 0.796,~SE = 0.041$) and predicting with the two-stage framework ($M = 0.820,~SE = 0.043$). This result supports hypothesis \textbf{H2}.


\textbf{Interpretation of Results.} We observe that clustering based on event sequences outperforms clustering based on primary actions. The baseline performs poorly after timestep $t=12$ (\fig{fig:fullseq}) as most users switch to shelf connections; there, the order of primary actions (which shelf to connect) is different among users, but they require the same secondary actions (bringing a shelf) and thus belong to the same event. The drops in accuracy for our method are caused by errors in prediction for participants that did not belong to any cluster, e.g. users 3, 6, 7 and 8 in Fig.~\ref{fig:event_sequences}, as well as by ``branching'' points, where participants that had identical sequences started to differentiate. For instance, participants 14 and 15 differentiate at $t=12$, where performance drops for all methods.

Similarly, we observe in Fig.~\ref{fig:twostage} that the two-stage clustering and clustering based on events only perform similarly up to $t=18$, from where users differentiate on how they connect the shelves to the boards, which is where the low-level preferences benefit prediction.


\section{Online Assembly Experiment}\label{sec:online}
\begin{figure}[hbt!]
\centering
\vspace{-1 ex}
\includegraphics[width=0.415\linewidth]{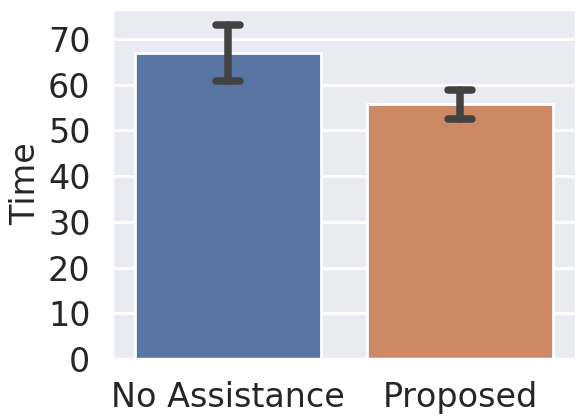}
\quad
\includegraphics[width=0.45\linewidth]{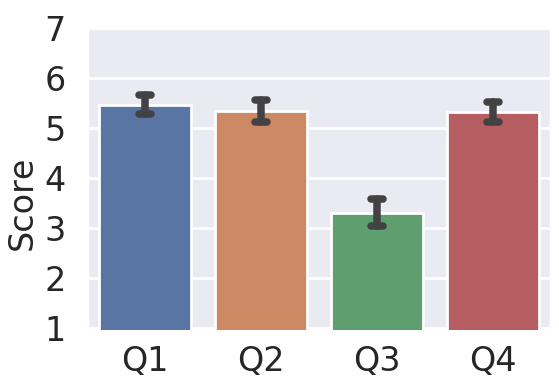}
\vspace{-0.5 ex}
\caption{(Left) Average time taken by users to complete the online game with and without assistance. (Right) Subjective responses of users for Q1: Assistance was according to your preference?, Q2: Assistance was helpful? Q3: Assistance was detrimental? Q4: Assistance reduced your effort?}
\label{fig:online_experiment}
\vspace{-1.5ex}
\end{figure}

We wish to show that predicting the secondary action to assist a new user improves the efficiency of the task. Therefore, we implemented an online shelf assembly game (see supplemental video). We first conducted an online study where $20$ users demonstrated their preference for playing the game and used that to learn the dominant high and low level preferences. As the game was simple, users had one dominant high-level preference and two low-level preferences on how to connect the shelves, which interestingly matched closely with how users clustered when connecting shelves in the real world (Fig.~\ref{fig:low-clusters}). We then conducted an online experiment with $80$ participants recruited through Amazon Mechanical Turk, where users played a shelf assembly game three times each for with and without assistance. The order of assistance versus no assistance was counterbalanced to reduce any learning effects. 

\textbf{H3.} We hypothesize that providing assistance to users by predicting the next secondary action improves the efficiency of the task, as compared to providing no assistance.

We compare the average time taken by $52$ (out of $80$) users who completed all game trials and survey questions, when playing with and without assistance using our two-stage inference method (see \fig{fig:online_experiment}). A paired t-test showed a statistically significant difference ($t(51) = 2.155$, $p = 0.036$) in the time required to assemble with ($M = 55.794,~SE = 3.439$) without assistance ($M = 66.948,~SE = 6.05$). Moreover in the subjective responses the users gave a high rating when asked if the assistance was according to their preference (Q1), was helpful (Q2) and reduced their effort (Q4). Accordingly they gave a low rating when asked if the assistance made their task more difficult (Q3). This informs us that performing the secondary actions as per our proposed method can indeed be helpful for the users.

\section{Robot-Assisted Ikea Assembly}\label{sec:robot}


Lastly we show how our proposed method can be used in a human-robot collaborative assembly setting. We demonstrate the online execution phase in the IKEA assembly task with two participants\footnote{We were unable to perform a complete human subjects experiment because of COVID-19 restrictions.}, who were instructed to perform the task in two different ways. We used the same setup as shown in \fig{fig:study} with a Kinova Gen 2 robot arm in the storage area.

The human-robot collaboration followed a turn-taking model where the user and the robot alternated in performing one primary action and one set of secondary actions, respectively. Therefore, the participant always stayed inside the workcell and performed all the connections, while the robot brought the required parts from the storage area to the workcell (Fig.~\ref{fig:user_preferences}). The experimenter inputted each primary action performed by the participant into the system, and the system inferred the subject's preference cluster and selected the next secondary action for the robot to perform.

The supplemental video shows the demonstration of the system. We observed that the robot assistance allowed the user to stay inside the workcell in a comfortable position. We hypothesize that this can be beneficial, not just in improving the efficiency of the task but also in reducing the human effort, and we plan to explore this in future work. 

\section{Conclusion}\label{sec:con}
We propose a two-stage clustering approach, inspired by clickstream analysis techniques, to identify the dominant preferences of users at different resolutions in a complex IKEA assembly task. We show how this approach can enable prediction of the next action required of the robotic assistant and how it can improve the efficiency of task execution.


A limitation of our work is that we do not model the effect of robot actions on the user~\cite{sadigh2016planning,nikolaidis2017human}, or the utility of actions with respect to team performance. Future work can also consider the confidence in the prediction, to decide if a robot should request additional information from the user. 

While we have focused on predicting the actions of a new worker on the same task that we have demonstrations for, our insights can be applied to the problem of having the same worker perform a new, slightly different task. Even though the order of events may be different, and some events may be task-specific, our system can learn the user's preference(s) on the events that are shared among the two tasks and proactively assist the user for these events. We are excited about demonstrating the applicability of our approach to this problem in future work. 




\bibliographystyle{IEEEtran}
\bibliography{references}

\end{document}